\documentclass{article}
\usepackage{preamble}
\usepackage{booktabs}
\title{Detection Accuracy for Evaluating Compositional Explanations of Units}

\author{%
    Sayo M.~Makinwa, Biagio La Rosa\thanks{Contact Author}
    , Roberto Capobianco\textsuperscript{1,2}\\
    \textsuperscript{1}Sapienza University of Rome\\
    \textsuperscript{2}Sony AI\\
    \texttt{sayomichaelmakinwa@gmail.com, \{larosa, capobianco\}@diag.uniroma1.it} \\
}

\begin{document}

\maketitle

\begin{abstract}
The recent success of deep learning models in solving complex problems and in different domains has increased interest in understanding what they learn. Therefore, different approaches have been employed to explain these models, one of which uses human-understandable concepts as explanations. Two examples of methods that use this approach are Network Dissection~\cite{bau2017} and Compositional explanations~\cite{mu2020}.
The former explains units using atomic concepts, while the latter makes explanations more expressive, replacing atomic concepts with logical forms. While intuitively, logical forms are more informative than atomic concepts, it is not clear how to quantify this improvement, and their evaluation is often based on the same metric that is optimized during the search-process and on the usage of hyper-parameters to be tuned.
In this paper, we propose to use as evaluation metric the Detection Accuracy, which measures units' \textit{consistency of detection} of their assigned explanations.
We show that this metric
(1) evaluates explanations of different lengths effectively,
(2) can be used as a stopping criterion for the compositional explanation search, eliminating the explanation length hyper-parameter, and 
(3) exposes new specialized units whose length 1 explanations are the perceptual abstractions of their longer explanations. The code is available at \url{github.com/KRLGroup/detacc-compexp}.
\\ \\
Explainable AI $\cdot$ Explainability $\cdot$ Deep Learning $\cdot$ Machine Learning $\cdot$ Metrics

\end{abstract}

\section{Introduction}

In the last decade, the interest on explaining what \textit{complex} deep learning models learn has grown due to the success of these models and the will to apply them on critical domains, like healthcare, where their decision process could improve our daily life.
Recent works show that units of models trained for a variety of tasks learn to detect human-understandable concepts, despite the fact that they are not trained to do it~\cite{gonzalez2016,vondrick2016,zhou2015}. This observation allows researchers to propose concept-based methods to provide explanations about the learned behavior of these models~\cite{kim2018,ghorbani2019,bau2017,mu2020}. An example of such a work is the framework of Network Dissection proposed by \citet{bau2017}. Starting from the assumption that explanations should be understandable to humans~\cite{miller2017a,molnar2019,slugoski1993}, it proposes to associate a human-understandable concept to each unit using a probing dataset.
Noting that assigning a single concept to each unit may be too simplistic to properly capture the units behavior, \citet{mu2020} extend the framework associating logical forms of concepts to units, calling these forms \textit{compositional explanations}.

\begin{figure}[t]
    \centering
    \includegraphics[scale=0.57]{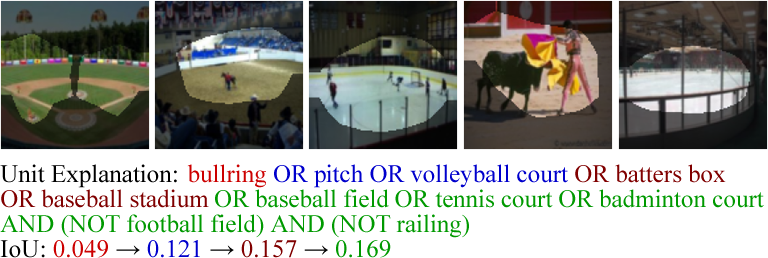}
    \caption{An example of an increasing length compositional explanation for a unit, and the evaluations for each explanation length.}
    \label{fig:comp_example}
\end{figure}

In both Network Dissection and compositional explanations, the same metric is used to generate explanations and to evaluate them (i.e. Intersection over Union score). This means that by design, 
the compositional explanations always score higher,
because the score increases with the number of concepts in the explanation (Figure \ref{fig:comp_example}). 
Hence, longer explanations are always deemed better than more precise ones. This is in contrast to the literature on explanations, which state that human-understandable explanations should be selected and precise \cite{hilton1996,miller2017a,mittelstadt2018,molnar2019,ylikoski2013}. Our hypothesis is that we need a different metric to compare concept-based methods and to properly assess when units are better explained by a method or another.

We propose to use Detection Accuracy for this purpose, which measures the 

\textit{consistency of detection}
of the explanations by the concerned units. 
We test the metric comparing Network Dissection and compositional explanation, obtaining a more balanced evaluation. Additionally, we show that Detection accuracy

\begin{enumerate}
    \item holds more information about the association between the concepts learned by units and the predictions of the model;
    
    \item can replace the maximum length hyper-parameter in the compositional search to generate variable length unit-optimal explanations;
    
    \item reveals new specialized units whose length 1 explanations are the perceptual abstractions of their longer explanations.
\end{enumerate}

The remaining sections are organized as follows; we briefly discuss existing work in concept explanations and evaluations in section 2, we present our approach in section 3 and then discuss our experiments and results in sections 4. In section 5, we present the conclusion and future work.

\section{Related Work}
\textbf{Explanation Methods.}\ \ \ \  Deep networks are complex models for which various class of explanation methods have been developed \cite{alejandro2020,dosilovic2018,xu2019}.
One such class of methods deals with the visualization of units, either by finding the most activated input image \cite{dalvi2019,zeiler2013,zhou2015},
by synthesizing an input using generative approaches \cite{le2011,nguyen2016},
or by generating saliency maps from gradients \cite{mahendran2014,simonyan2014}.
The extracted concepts by these methods can be exploited in several ways to obtain explanations.
\citet{kim2018} propose to generate concept-based explanations by computing directional derivatives on a linear classifier to measure the importance of a concept from a user-defined dataset, to a class prediction. \citet{ghorbani2019} suggests to segment concepts, group these segmentations, and evaluate them against classes to know which concepts are more important for which class. These two approaches explain the \textit{local} behaviour around a particular prediction.
Conversely, \citet{bau2017} propose to generate \textit{global} explanations by explaining each unit in the model, using a framework called Network Dissection. This method extracts unit activation above a threshold on a large dataset, and computes the overlap with the concepts in the dataset. \citet{mu2020} then extends Network Dissection by producing explanations that are logical compositions of concepts from the dataset.

\vspace{3mm}

\textbf{Evaluation of Explanations.}\ \ \ \  Extensive work has been done to establish general guidelines for assessing explanations \cite{hilton1996,miller2017a,rosenfeld2021,yang2019,ylikoski2013}, which has now formed a basis for evaluation of explanations and explanation methods.
The majority of the current approaches measure how generated explanations change in response to perturbations in the input data, either as a result of completely removing a feature \cite{ancona2017,lundberg2017}, or just some modifications of the feature \cite{lundberg2017,samek2017,sundararajan2017,yeh2019b}. 
Alternative approaches are the sanity checks~\cite{adebayo2018} that randomize the weights of the layers and inspect how the generated explanations change, or the method of \citet{lin2020}, which create a backdoor in the model, adding a trigger on the data for a specific prediction, and observing how explanations detect this attack.
Finally, \citet{kim2016} propose to evaluate example-based explanations by generating criticisms to the explanations, in addition to the positive examples.
\citet{kim2018} is the closest to our work; it explains a class prediction of a model by measuring how consistently a concept influences the class prediction, using a user-defined concept dataset. It is however local to a particular prediction, and it is also embedded within an explanation method. 
 
Our work evaluates the global relationship of concept explanations to the units they explain,
does not require users to build and annotate their own concept dataset, and is separate from the explanation method.

\section{Methodology}
This section describes the Network Dissection framework and the procedure to generate compositional explanations, and then
the proposed explanation evaluation metric, Detection Accuracy. 

\subsection{Network Dissection and Compositional Explanations}
\label{sec:comp} 
Network Dissection \cite{bau2017} 
explains a unit by measuring the alignment between the unit and semantic concepts from a probing dataset.
First, we collect the unit's activation $A_u(X)$ on the images in the dataset and compute its distribution $a_u$, then we determine the top quantile level $T_u$ for this unit such that $P(a_u > T_u) = 0.005$. 
We then scale $A_u(X)$ up to the dimension of concepts' annotation mask and convert $A_u(X)$ to a binary form where $M_u(X) \equiv A_u(X) \geq T_u$.
Finally, we compute the Intersection over Union (IoU) score as follows:
\begin{equation}
    \newcommand*{\iounumer}{\sum \lvert M_u(X) \cap G_E(X) \rvert}
    \newcommand*{\ioudenom}{\sum \lvert M_u(X) \cup G_E(X) \rvert}
    IoU_{u, E} = \frac{\iounumer}{\ioudenom}
    \label{eq:iou}
\end{equation}
where $G_E(X)$ is the binary annotation mask for a concept explanation $E$, and $\lvert \:.\: \rvert$ is the set cardinality. 
We compute the IoU score for each explanation and assign the highest scoring explanation to the unit:

\begin{equation}
    \overline{E}_u = \operatorname*{argmax}_{E \in \mathcal{L(C)}^1} IoU_{u,E}
    \label{eq:iou_search}
\end{equation}
where 
$\mathcal{L(C)}^1$ is the set of all the atomic concepts in the dataset.

Compositional explanations \cite{mu2020} replace the set of atomic concepts
$\mathcal{L(C)}^1$ with a set of $n$-ary logical forms
$\mathcal{L(C)}^n$ 
composed of up to length $n$ concepts.
We build the set of logical forms incrementally using \textit{beam search}, which 
takes as input the set of atomic concepts, computes their IoU, and initializes the beam with the 
explanations with the top $B$ IoU.
$B$ is the beam size. 
It then combines each explanation in the beam with each atomic concept, it merges the resulting set with the set in the beam to create the next length set of logical forms, and computes the IoU of all the explanations in this new set, keeping only the explanations with the top $B$ IoU in the beam. 
The search returns the explanation with the highest IoU as the explanation for the unit:
\begin{equation}
    \overline{E}_u = \operatorname*{argmax}_{E \in \mathcal{L(C)}^n_B} IoU_{u,E}
    \label{eq:comp_search}
\end{equation}

\subsection{Detection Accuracy} 
The intuition behind the Detection Accuracy score is that an explanation is associated to a unit to the degree of how consistently the unit \textit{detects} it. 
To measure the Detection Accuracy score of a unit $u$ on an explanation $E$ that is composed of $n$ concepts $c_{1-n}$ over a dataset $X$ of images, we evaluate the following equation: 
\begin{equation} 
    \newcommand*{\detaccnumer}{\sum_{imgs} M_u(X) \cap G_E(X)}
    \newcommand*{\detaccdenom}{\sum_{imgs} G_E(X)}
    DetAcc_{u,E} = \frac{\detaccnumer}{\detaccdenom}
    \label{eq:detacc}
\end{equation}
where $G_E$ is the gold binary mask of explanation $E$ 
in the dataset, and $M_u$ is unit $u$'s binary activation map described in section \ref{sec:comp}.
We compute $M_u(X) \cap G_E(X)$ over binary pixels while $\sum_{imgs}$ represents an image-wise summation. 
In particular, for every image in the dataset that contains the given explanation, we compare the pixels of the image to the pixels of the unit activation on the image via a pixel-wise binary operation, to find out if the unit activates on the image and if this activation overlaps the explanation. After this, we count the total number of images where the unit activates on the explanation and compute the ratio of this number to the number of images where the explanation exists.
The idea is to measure how consistently the unit \textit{sees} the explanation, 
not minding how well it covers the explanation each time.
A high Detection Accuracy score of a unit on an explanation corresponds to a high consistency in the unit's detection of the explanation.
Therefore, if multiple explanations are given for a unit, we consider the explanation with the highest Detection Accuracy score as the one that better explains the unit.

\section{Experiments}
This section describes the dataset and models used for our experiments. Then, it presents and analyzes the results on the quality of Detection Accuracy's evaluations, showing its properties.

\subsection{Setup}

As dataset, we use the ADE20K \cite{zhou2016} scene parsing dataset. It contains 22,210 densely annotated images on the pixel level with human-understandable concepts from the Broden Dataset\cite{bau2017}, categorized into classes of scenes (468), colors (11), parts (96), and objects (518). We consider only concepts with at least 5 samples, leaving only 1093 of the 1,105 concepts.

Following \citet{mu2020}, the experiments are conducted on the units of the final layer of ResNet-18 \cite{he2016}, AlexNet \cite{krizhevsky2012}, DenseNet-161 \cite{huang2016}, and ResNet-50 \cite{he2016}, trained on the Places365 dataset \cite{zhou2018}.

\subsection{Detection Accuracy as Evaluation Metric}
\begin{figure}[t]
    \centering
    \includegraphics[scale=0.19]{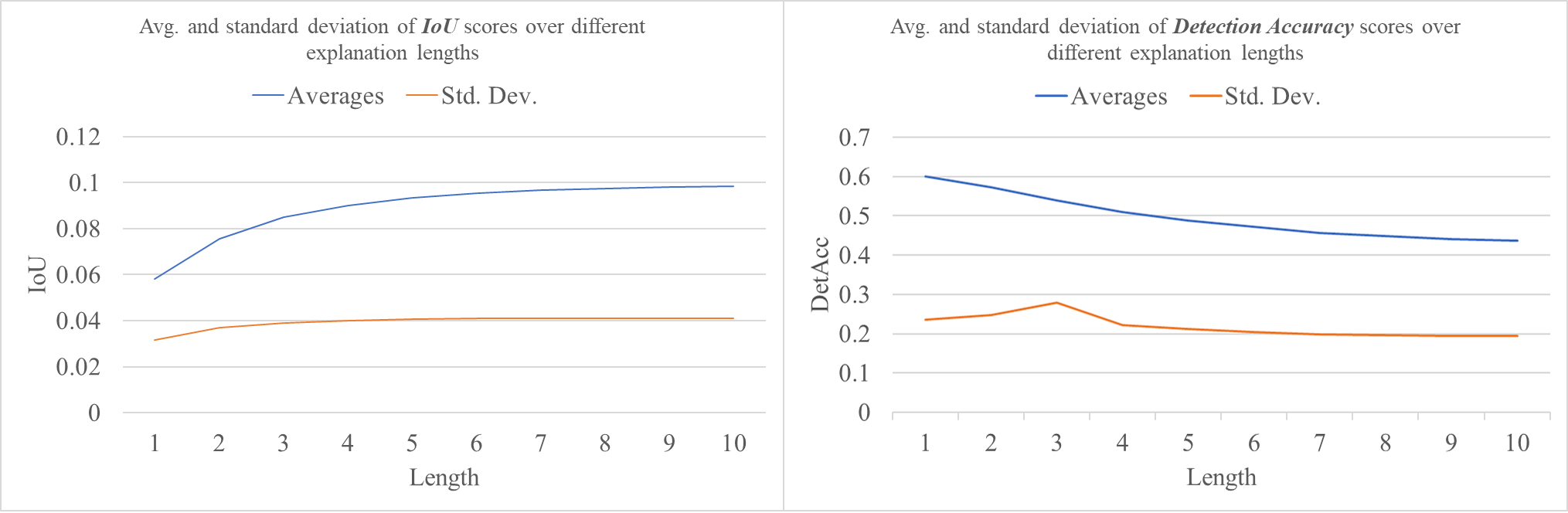}
    \caption{
    Plots of averages and standard deviation of IoU (L) and Detection Accuracy (R) scores over all the units of ResNet-18 for explanation lengths 1 to 10. The plots show how the IoU scores generally increase as explanation length increases, while the Detection Accuracy scores generally reduce.
    }
    \label{fig:scores_avgs}
\end{figure}

First, we test Detection Accuracy as an evaluation metric against the IoU, comparing their evaluation on Network Dissection and compositional explanations. We fix the length of compositional explanations to 3 as in \citet{mu2020}.
The results (Table \ref{tab:eval_res}) show that the IoU scores of the compositional explanations are higher for nearly 100\% of the units across different models. 
The Detection Accuracy's evaluation on the other hand is more distributed, rating the Network Dissection explanations higher for approximately half the number of units, and the compositional explanations higher for the other units across all the models.
To observe the behaviour of the scores as maximum explanation length increases, we further generate explanations of lengths up to 10.
We see that the Detection Accuracy penalizes verbose explanations while the IoU rewards them (Figure \ref{fig:scores_avgs}), which establishes that when Detection Accuracy scores a longer explanation higher, it has merely rated the actual value of the explanation rather than rewarding the length.


\begin{table}[b]
\centering
\caption{Evaluation results of IoU and Detection Accuracy on Network Dissection explanations and length 3 Composition explanations for all the units in ResNet-18, AlexNet, ResNet-50, and DenseNet-161 models. Each percentage represents the percentage of units in the model that have the corresponding \textit{higher} score for the explanation. Note that the scores do not necessarily add up to 100\% for each model because a few compositional explanations have exactly the same scores.
}
\begin{tabular}{rcccccccc}
    \toprule
    & \multicolumn{2}{c}{\begin{tabular}[c]{@{}c@{}}ResNet-18\\ (512 units)\end{tabular}}  & \multicolumn{2}{c}{\begin{tabular}[c]{@{}c@{}}AlexNet\\ (256 units)\end{tabular}} & \multicolumn{2}{c}{\begin{tabular}[c]{@{}c@{}}ResNet-50\\ (2048 units)\end{tabular}} & \multicolumn{2}{c}{\begin{tabular}[c]{@{}c@{}}DenseNet-161\\ (2208 units)\end{tabular}} \\
    \midrule
    & NetDis. & Comp. & NetDis. & Comp. & NetDis. & Comp. & NetDis. & Comp. \\ \midrule
    IoU & 0.2\% & 99.8\% & 0\% & 100\% & 0.05\% & 99.95\% & 0.05\% & 99.95\% \\ 
    DetAcc & 56.45\% & 43.16\% & 37.89\% & 60.94\% & 53.76\% & 45.31\% & 56.46\% & 42.82\% \\
    \bottomrule
\end{tabular}
\label{tab:eval_res}
\end{table}

\begin{table}[b]
\centering
\caption{Correlation scores of all units in the last layer of the ResNet-18 model over images associated to each unit and its Network Dissection or compositional explanations. The best scores are marked in bold.
}
\begin{tabular}{rccccc}
    \toprule
    & \multicolumn{2}{c}{Union Corr.} & & \multicolumn{2}{c}{Intersection Corr.} \\
    \midrule
    & IoU & DetAcc & & IoU & DetAcc \\ 
    \midrule
    NetDissect & 0.2224 & \textbf{0.3615} & & 0.0365 & \textbf{0.5799} \\ 
    Comp. & 0.2733 & \textbf{0.3294} & & 0.1158 & \textbf{0.4876} \\ 
    \bottomrule
\end{tabular}
\label{tab:correlation}
\end{table}

Proceeding to inspect how valuable the Detection Accuracy evaluation is, we measure the correlation between the IoU score and the prediction accuracy of the model, and the correlation between the Detection Accuracy score and the prediction accuracy of the model.
To do this, we take each unit and its explanation, we select images from the dataset on which the unit fires and those on which the explanation is present, then we combine both sets by finding the union and the intersection to create two new sets.
We then measure the model prediction accuracy on these sets and compute correlation of these accuracy scores with the IoU and Detection Accuracy of the explanations.
Results show that the Detection Accuracy positively correlates to model prediction accuracy on both Network Dissection and compositional explanations, and these correlation scores are higher the than IoU's correlation scores (Table \ref{tab:correlation}).
This means that the Detection Accuracy scores better reflect how the concepts in the input images contribute to the correctness of the model predictions.

\begin{figure}[t]
    \centering
    \includegraphics[scale=0.39]{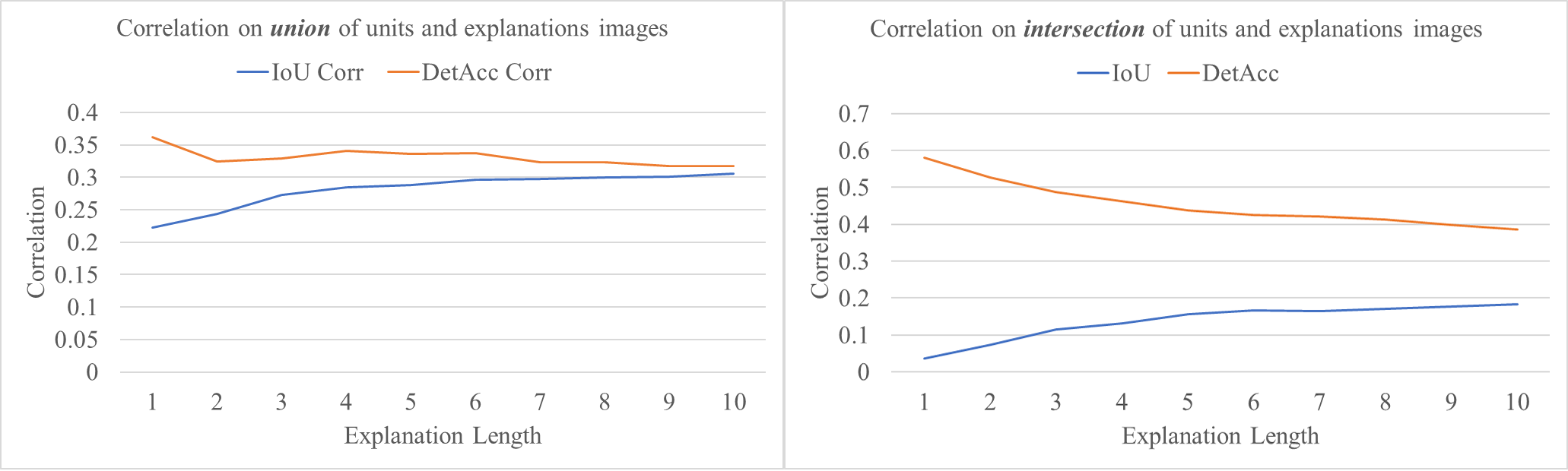}
    \caption{Plots of IoU and Detection Accuracy correlation scores for explanation lengths 1 to 10 over all the units in the last layer of ResNet-18 model, computed over the \textit{union} [L] and \textit{intersection} [R] of images associated with each unit and its explanation. The plots show how the Detection Accuracy score is consistently more correlated to model accuracy than the IoU score. We also see how the IoU correlation scores increase with explanation length.
    }
    \label{fig:correlation}
\end{figure}

Additionally, we can observe that the correlation for Network Dissection explanations are higher for the Detection Accuracy, while the correlation for compositional explanations are higher for the IoU.
This suggests a behaviour where the Detection Accuracy's correlation prefers precise explanations, while the IoU's correlation prefers verbose explanations. We test this further by computing correlation for explanations of lengths up to 10, since \citet{mu2020} showed that the increase in the IoU of explanation lengths greater than 10 is not substantial.
Results show that the Detection Accuracy's correlation scores generally decrease as explanation length increases, while that of the IoU generally increases (Figure \ref{fig:correlation}).
From the Detection Accuracy's perspective, the precise explanations sufficiently capture the concepts that are important to the model predictions, while the extra concepts added in the verbose explanations rewarded by IoU do not have the same level of importance.

\subsection{Detection Accuracy as Optimization Metric}

\label{sec:select_stop}

In this test, we investigate if it is possible to use the Detection Accuracy also for optimizing explanations. First, we insert Detection Accuracy into the beam search by selecting the explanation with the highest Detection Accuracy from the beam at each step of the search. Note that since the beam has a size $B$, the explanations in the beam are ones with the top IoU scores. The best explanation at each step is therefore the one with the best Detection Accuracy from the set of top $B$ IoU explanations. Equation \ref{eq:comp_search} then becomes:
\begin{equation}
    \overline{E}_u = \operatorname*{argmax}_{E \in \mathcal{L(C)}^n_B} DetAcc_{u,E}
    \label{eq:comp_select_stop}
\end{equation}
In addition to using Detection Accuracy to select the best explanation from the beam, we also allow the search to continue to run until the Detection Accuracy score for the unit can no longer be improved.

\begin{figure}[t!]
    \centering
    \includegraphics[scale=0.38]{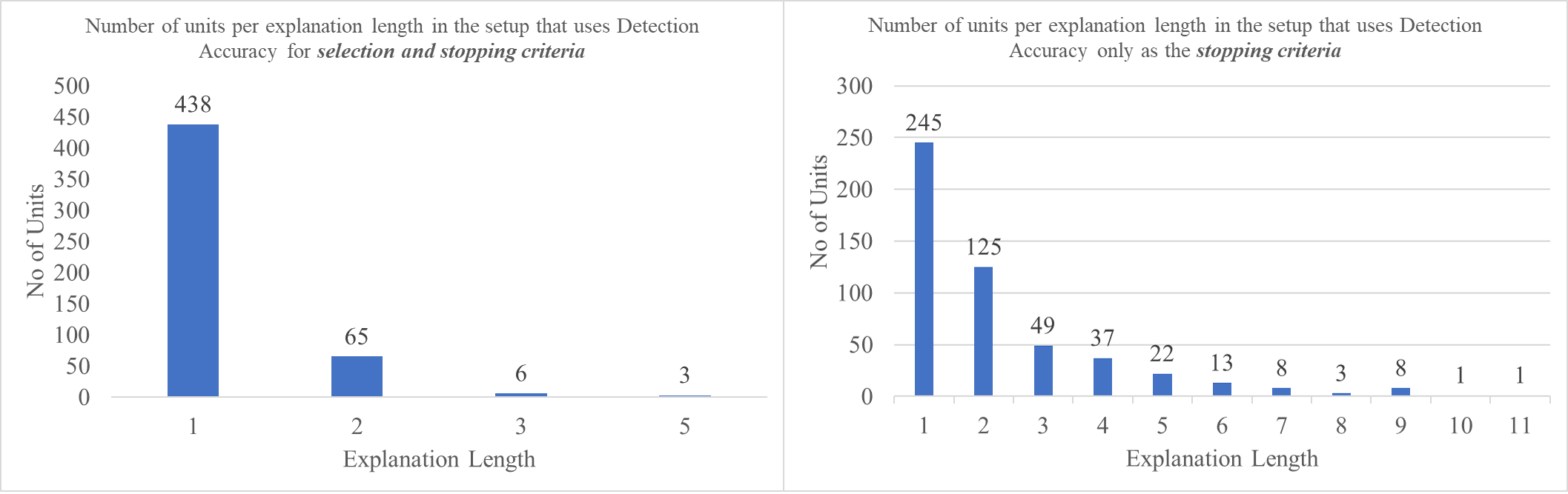}
    \caption{Summary of explanation lengths across units of the last layer of ResNet-18 model from two modified compositional methods; where the best explanations are chosen from the beam using their Detection Accuracy scores [L], and where Detection Accuracy is used only as a stopping criterion [R]. The setup in [L] returns mostly length 1 explanations for units, while the lengths of the explanations returned for units by the setup in [R] are more spread.
    }
    \label{fig:select_stop_units_dist}
\end{figure}

Figure \ref{fig:select_stop_units_dist} shows that $\sim$85.5\% of the units (438 of 512) are assigned explanations of length 1. This is because the most frequently detected explanations are greedily selected early in the search. 
We then proceed to 
inspect how valuable the explanations are
using the correlation of the model accuracy to their IoU and Detection Accuracy scores. Both correlation scores dropped compared to the regular compositional explanations method (Table \ref{tab:select_stop_corr}). This means that the concepts in these explanations have a weaker relationship to the model predictions and therefore, the Detection Accuracy cannot be exploited in generating explanations as described in this experiment.

\begin{table}[b!]
\centering
\caption{Correlation scores of the regular compositional method and the modified compositional method, where the best explanations are \textit{selected} from the beam using their Detection Accuracy scores. }
\begin{tabular}{rccccc}
    \toprule
    & \multicolumn{2}{c}{Union Corr.} & & \multicolumn{2}{c}{Intersection Corr.} \\
    \midrule
    & IoU & DetAcc & & IoU & DetAcc \\ 
    \midrule
    Len3 Comp. Expl. & 0.2733 & 0.3294 &  & 0.1158 & 0.4876 \\
    Comp.+Select & 0.2341 & 0.2247 &  & 0.0368 & 0.3603 \\ 
    \bottomrule
\end{tabular}
\label{tab:select_stop_corr}
\end{table}

\label{sec:stopping}

\begin{figure}[t!]
    \centering
    \includegraphics[scale=0.59]{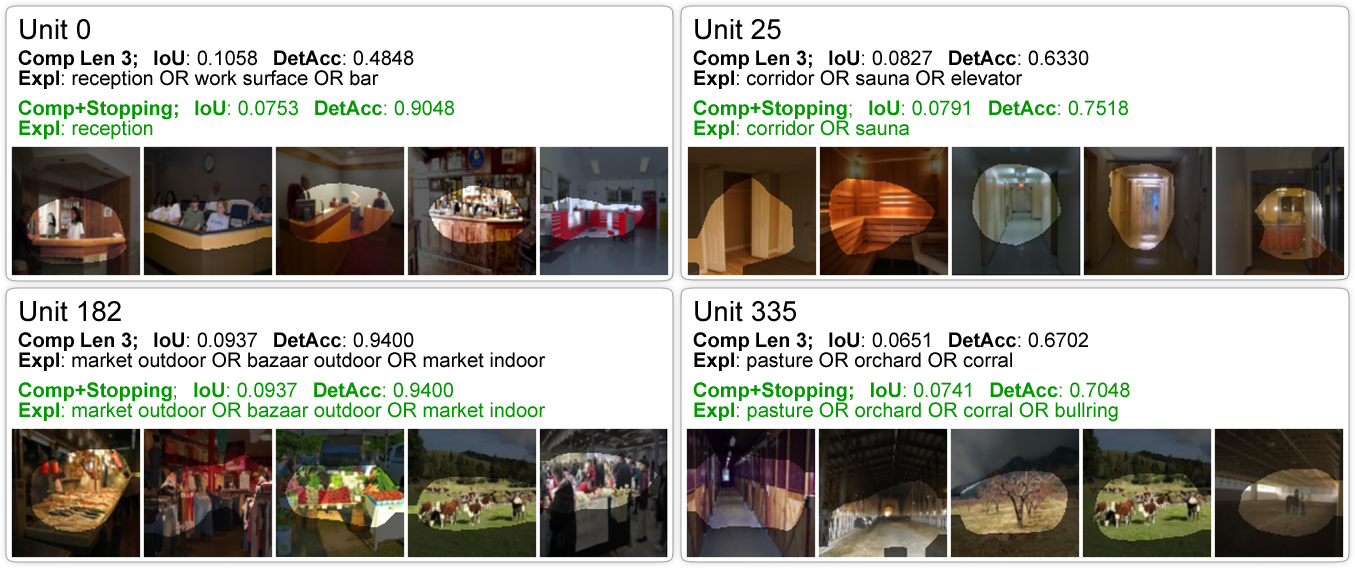}
    \caption{Examples of explanations from the modified compositional explanations method where Detection Accuracy is used \textit{only} to stop the search. The explanations are compared to fixed length compositional explanations.
    }
    \label{fig:AB_ex}
\end{figure}

\begin{table}[b]
\centering
\caption{Comparison of correlation scores among the compositional methods on the ResNet-18 model.
}
\begin{tabular}{rccccccccc}
    \toprule
    & & & \multicolumn{2}{c}{Union Corr.} & & & & \multicolumn{2}{c}{Intersection Corr.} \\
    \midrule
    & & & IoU & DetAcc & & & & IoU & DetAcc \\ 
    \midrule
    Len3 Comp. Expl. & & & 0.2733 & 0.3294 & & & & 0.1158  & 0.4876 \\
    Comp.+Select & & & 0.2341 & 0.2247 & & & & 0.0368 & 0.3603 \\
    Comp.+Stopping & & & 0.2007 & 0.3409 & & & & -0.0546 & 0.5331 \\ 
    \bottomrule
\end{tabular}
\label{tab:stopping_corr}
\end{table}
At this point we question if it is possible to use Detection Accuracy only to stop the beam search, removing the maximum explanation length hyper-parameter and letting the search run until the best Detection Accuracy score so far is no longer improved. 
In this case, we find that the explanation lengths become more varying than before, while not losing the preference for precision in the explanations (Figure \ref{fig:select_stop_units_dist}).
In probing how valuable these new explanations are, we also compute the correlation of the model accuracy to the IoU and Detection Accuracy scores (Table \ref{tab:stopping_corr}).
We find that for the Detection Accuracy, not only are the correlation scores better than the scores from the previous experiment, they also improve on the correlation scores from the regular compositional method.
The IoU correlation scores however did not improve, further establishing that the IoU prefers longer explanations.
We argue that these varying length explanations allow units to be described with explanations of lengths that are optimal for them, as opposed to \textit{forcing} them shorter or longer like in the regular setup with the maximum explanation length hyper-parameter (See examples in Figure \ref{fig:AB_ex}).

\begin{figure}[t]
    \centering
    \includegraphics[scale=0.35]{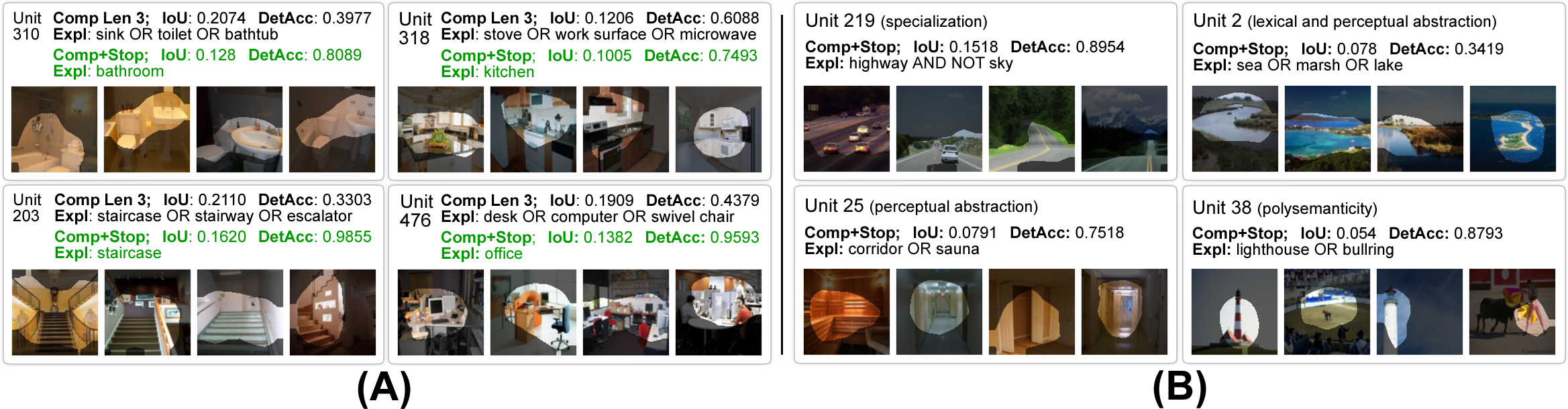}
    \caption{Examples of the different kinds of units in the compositional setup where Detection Accuracy is used only to stop the search. (A) are examples where the these new explanations are perceptual abstractions of the regular compositional explanations, while (B) are examples of explanations of length greater than 1 and their groupings, similar to \citet{mu2020}.}
    \label{fig:AB_ex2}
\end{figure}

\begin{figure}[t]
    \centering
    \includegraphics[scale=0.58]{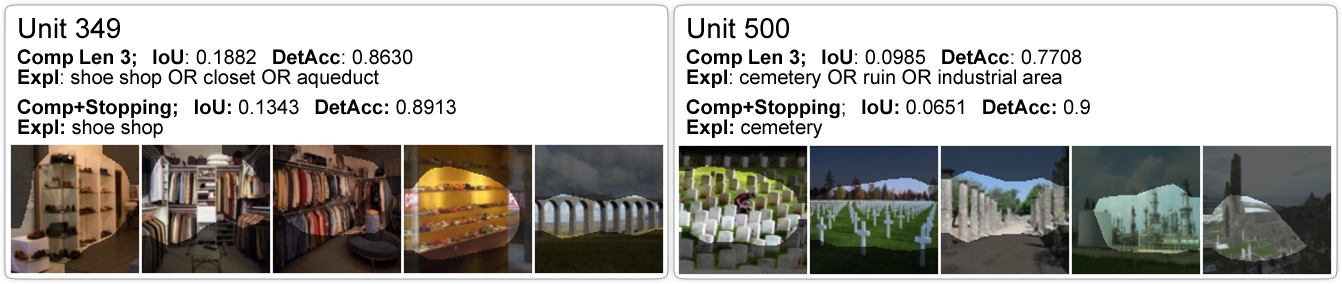}
    \caption{Examples of exceptions to the examples described in Figure \ref{fig:AB_ex2} (A).}
    \label{fig:specialized_alt_ex}
\end{figure}

\vspace{3mm}
\textbf{New Specialized Units.}\ \ \ \ 
Querying further the results from the last setup where we use the Detection Accuracy only as a stopping criterion, we find that the Detection Accuracy score exposes a new set of specialized units, keeping their explanations to length 1. 
These new length 1 explanations correspond to the perceptual abstractions described by their \textit{longer} explanations from the regular compositional setup (Figure \ref{fig:AB_ex2} (A)). 
They are also different from the $L_1$ AND NOT $L_2$ form described in \citet{mu2020}.
It is important to note that not all the units with length 1 explanations in this setup are in this category. A manual inspection of the explanations of the units shows that from the 245 in 512 units with length 1 explanations, only 179 ($\sim$73\%) of them fall in this category. The other 66 units ($\sim$27\%) have explanations where the longer explanations failed to improve the Detection Accuracy during the explanation search (Figure \ref{fig:specialized_alt_ex}).
Additionally, units with explanation lengths greater than 1 in this setup have the same behaviour noted in \citet{mu2020}; they have either learned lexically coherent or incoherent perceptual abstractions, or they are polysemantic. See Figure \ref{fig:AB_ex2} (B) for examples.

\section{Conclusion and Future Work}
In this work, we proposed an evaluation of compositional explanations of CNN units using Detection Accuracy. 
By evaluating Network Dissection and compositional explanations of different lengths, we showed that Detection Accuracy's evaluation of explanations is more objective and that it encodes the relationship between the concepts units learn and the predictions of the model. 
We further presented a modified algorithm for compositional explanations where the maximum explanation length hyper-parameter is removed using Detection Accuracy. 
We established that the results from this modified procedure are a better reflection of what units learn, while also showing how a new set of specialized units are exposed. 
Finally, we showed how that Detection Accuracy is more valuable as an evaluation method than an explanation method.

Following these results, our suggestions for future work as are follows:

\begin{enumerate}
    \item Past work \cite{alqahtani2021,yeom2019} showed that pruning of CNNs can be informed using feature importance scores from explanations. 
    Can Detection Accuracy scores achieve a better pruning of units?
    
    \item In the setup of Detection Accuracy as stopping criterion for the compositional search,
    we noted that the setup strictly sticks to the shorter explanation if the Detection Accuracy of the next length explanation is not better. Would introducing an adaptive or decaying tolerance term produce better results?
    
\end{enumerate}

\bibliography{bibliography}

\end{document}